\def\year{2017}\relax
\documentclass[letterpaper]{article} 
\usepackage{aaai18}  
\usepackage{times}  
\usepackage{helvet}  
\usepackage{courier}  
\usepackage{url}  
\usepackage{graphicx}  
\usepackage{subcaption}
\usepackage{amssymb,amsmath}
\usepackage{url}

\usepackage{color}
\usepackage{ifthen}
\usepackage{newclude}
\usepackage{epstopdf}
\usepackage{flushend}
\usepackage{xspace}
\usepackage{graphicx}
\usepackage{tikz}
\usepackage{pgfplots, pgfplotstable}
\usepgfplotslibrary{statistics}
\usepackage{xspace}
\usepackage{multirow}
\usepackage{tabularx}

\makeatletter
\def\url@foostyle{%
  \@ifundefined{selectfont}{\def\UrlFont{\rm}}{\def\UrlFont{\rmfamily}}}
\makeatother
\usepackage[hyphenbreaks]{breakurl}
\usepackage[hang,flushmargin]{footmisc}

\definecolor{heraldGreen}{rgb}{0.0,0.4,0.0}

\newcommand{\dswebsci}{{Cyberbullying}\xspace}
\newcommand{\dswsdm}{{Sarcasm}\xspace}
\newcommand{\dsicwsm}{{Hate}\xspace}
\newcommand{\dsnaacl}{{Offensive}\xspace}

\newcommand{\websci}{{cyberbullying}\xspace}
\newcommand{\wsdm}{{sarcasm}\xspace}
\newcommand{\icwsm}{{hate}\xspace}
\newcommand{\naacl}{{offensive}\xspace}

\makeatletter
\def\url@leostyle{%
	\@ifundefined{selectfont}{\def\UrlFont{}}%
	{\def\UrlFont{}}%
}
\makeatother
\usepackage[hyphenbreaks]{breakurl}

\begin{document}

\sloppy

\title{A Unified Deep Learning Architecture for Abuse Detection}

\author{Antigoni-Maria Founta$^\ddagger$,
Despoina Chatzakou$^\ddagger$,
Nicolas Kourtellis$^\sharp$,\\
{\Large \bf Jeremy Blackburn$^\dagger$,
Athina Vakali$^\ddagger$,
Ilias Leontiadis$^\sharp$}\\
$^\ddagger$Aristotle University of Thessaloniki \hspace{0.2cm}
$^\sharp$Telefonica Research \hspace{0.2cm}
$^\dagger$University of Alabama\\ {\{founanti,deppych,avakali\}@csd.auth.gr}, 
{\{nicolas.kourtellis,ilias.leontiadis\}@telefonica.com},\\ {blackburn@uab.edu}}

\maketitle

\begin{abstract}
Hate speech, offensive language, sexism, racism and other types of abusive behavior have become a common phenomenon in many online social media platforms. In recent years, such diverse abusive behaviors have been manifesting with increased frequency and levels of intensity. This is due to the openness and willingness of popular media platforms, such as Twitter and Facebook, to host content of sensitive or controversial topics. However, these platforms have not adequately addressed the problem of online abusive behavior, and their responsiveness to the effective detection and blocking of such inappropriate behavior remains limited. In fact, up to now, they have entered an arms race with the perpetrators, who constantly change tactics to evade the detection algorithms deployed by these platforms. Such algorithms are typically custom-designed and tuned to detect only one specific type of abusive behavior, but usually miss other related behaviors.

In the present paper, we study this complex problem by following a more holistic approach, which considers the various aspects of  abusive behavior. To make the approach tangible, we focus on Twitter data and analyze user and textual properties from different angles of abusive posting behavior. We propose a deep learning architecture, which utilizes a wide variety of available metadata, and combines it with automatically-extracted hidden patterns within the text of the tweets, to detect multiple abusive behavioral norms which are highly inter-related. We apply this unified architecture in a seamless, transparent fashion to detect different types of abusive behavior (hate speech, sexism vs. racism, bullying, sarcasm, etc.) without the need for any tuning of the model architecture for each task. We test the proposed approach with multiple datasets addressing different and multiple abusive behaviors on Twitter. Our results demonstrate that it largely outperforms  the state-of-art methods (between 21 and 45\% improvement in AUC, depending on the dataset). 
\end{abstract}

\section{Introduction}\label{sec:intro}

Social media ubiquity has raised concerns about emerging problematic phenomena such as the intensity of  abusive behavior. Unfortunately, this problem is difficult to deal with since it has many ``faces'' and exhibits complex interactions among social media users. Such multifaceted abusive behavior involves instances of hate speech, offensive, sexist and racist language, aggression, cyberbullying, harassment, and trolling~\cite{waseem2017understanding,sanchez2011twitter}. Each form of abusive behavior has its own characteristics, and manifests differently, depending on the social media scope, the users participating in it, the topic's sensitivity,  and the particular platform it takes place on. Indeed, popular social media platforms like Twitter and Facebook are not immune to abusive behavior, even though they have devoted substantial resources to deal with it~\cite{salon}.

This type of behavior is harmful both socially, reducing the proclivity and trust of users to the particular online social media platform, as well as from a business perspective. For example, concerns about racist and sexist attacks on Twitter seem to have impacted a potential sale of the company.\footnote{Did trolls cost Twitter 3.5bn and its sale? \url{goo.gl/PlIL66}} It is apparent that social media platforms have not adequately addressed this problem so far. Instead, they iteratively announce new measures to curve abuse by adapting their detection mechanisms (e.g., Twitter 
\footnote{A Calendar of Our Safety Work (Twitter): \url{https://goo.gl/sX2Apb}}
\footnote{\url{https://twitter.com/jack/status/919028950650589184}}
).

The research community has also made attempts at detecting abusive behavior. For example, there have been various works attempting to detect hate speech
\cite{djuric2015hate},
\cite{warner2012detecting},
\cite{waseem2016hateful},
\cite{badjatiya2017deep},
cyberbullying 
\cite{chatzakou2017mean},
\cite{riadi2017detection},
\cite{dinakar2011modeling},
and even to address all kinds of abusive behavior
\cite{davidson2017automated},
\cite{chen2012detecting},
\cite{nobata2016abusive}.
Furthermore, various techniques have been applied to detect offensive language
\cite{xiang2012detecting},
\cite{clarke2017dimensions},
\cite{mehdad2016characters},
and even racism and sexism
\cite{kwok2013locate},
\cite{lozano2017requiem},
\cite{jha2017does}.
However, these solutions are typically custom built and tuned for a specific platform and \emph{type} of abusive behavior, and not generalizable.

Furthermore, abusive behavior cannot be assumed just by a ``monolithic'' consideration of the content (e.g., text of an individual post). Instead, in this work, we follow a more ``holistic'' approach to consider other facts such as the users' prior behavior, their social network, their popularity, their account settings, and even the metadata of posts, to reveal a more global abusive behavior tendency. We tackle the problem by designing a novel, \emph{unified} deep-learning architecture, able to digest and combine any available attributes, in order to detect abusive behavior. The deep-learning approach allows us to capture subtle, hidden commonalities and differences between the various abusive behaviors within the same model. Our method is a global and lightweight solution, with the capacity to take into account the plethora of available (meta)data, to recognize various types of abusive behavior, and without too much feature engineering and model tuning. We show that the combination of all available metadata with the proposed training methodology can substantially outperform the state of the art over a variety of datasets, each of which captures a different facet of abusive behavior: i)~\websci, ii)~\naacl, iii)~\icwsm, and iv)~\wsdm.

More concretely, we make the following contributions:~
\begin{itemize}
	\item We demonstrate the importance of combining all available (meta)data for detecting abusive behavior. We measure its importance by experimenting and producing superior results, with a deep-learning-based architecture able to combine the available input. To the best of our knowledge, we are the \emph{first to demonstrate the power of such a unified architecture in detecting various facets of abuse} in online social networks.
	
	\item We show how naive training methodology fails to make optimal use of heterogeneous inputs. To address this, we implement a training technique that focuses separately on each input by alternating training  between them. \emph{This optimization substantially boosts detection capabilities}, as it allows the model to avoid considering only the  most dominant features for each task.
	
	\item We show that our architecture is \emph{portable} across different forms of abusive behavior, as opposed to previous work which uses customized detectors for each type of abuse. We experiment on four datasets covering several forms of abuse, and find that our unified architecture works across all four \emph{without the need for any tuning or reconfiguring}. Our results shed light on how different feature types contribute to abuse detection, and provide evidence that text-only features alone are insufficient to reliably detect generic abusive behavior.
	
	\item We demonstrate that our methodology can be easily adapted for the detection of toxic behavior in domains such as online gaming, \emph{without further tuning}.
	
\end{itemize}

\section{Background and Related Work}\label{sec:back}

\textbf{Problem.}
Abuse detection has been an increasingly trending topic over the past few years. Numerous studies have been published, trying to address this problem especially in social networks, and in various forms.
Hate speech detection
\cite{djuric2015hate},
\cite{warner2012detecting},
\cite{waseem2016hateful},
\cite{badjatiya2017deep},
cyber-bullying identification
\cite{chatzakou2017mean},
\cite{riadi2017detection},
\cite{dinakar2011modeling},
and the detection of abusive
\cite{chen2012detecting},
\cite{nobata2016abusive},
\cite{davidson2017automated},
or offensive language
\cite{xiang2012detecting},
\cite{clarke2017dimensions},
\cite{mehdad2016characters}
are some of the facets of this problem.
In fact, some works try to detect more specific types of hateful behaviour, such as racism \cite{kwok2013locate}, \cite{lozano2017requiem}, or sexism \cite{jha2017does}. However, as \cite{waseem2017understanding} point out, there are many similarities between these subtasks, and scholars tend to group them under ``umbrella terms'' - like \cite{schmidt2017survey} do for hate speech - or use them interchangeably. Yet, major advancements on these tasks are quite new and many of the related studies are preliminary.

\textbf{Methods.}
Most of previous works use traditional machine learning classifiers, such as logistic regression \cite{xiang2012detecting}, \cite{waseem2016hateful},
\cite{davidson2017automated} and support vector machines \cite{warner2012detecting}, or ensemble classifiers of such traditional methods \cite{burnap2015cyber}. Some studies experiment with deep learning on this task, especially after the major advancements of the last years. Due to the large amount of related research concerning these tasks, we only analyze works that are most relevant with ours in terms of domain and methodology, like in \cite{badjatiya2017deep}, \cite{gamback2017using}, and \cite{park2017one}.

In the case of~\cite{badjatiya2017deep}, they focus on the problem of hate speech detection, and specifically attempt to detect racism and sexism, by applying various deep learning architectures. These architectures include Convolutional Neural Networks (CNNs), Long Short-Term Memory Networks (LSTMs), and FastText~\cite{joulin2016bag}, combined with numerous features like TF-IDF and Bag of Words (BoW) vectors. Their LSTM classifier with random embeddings seems to get significantly improved performance compared to baseline methods.

In \cite{gamback2017using} they also use deep learning models to address hate speech on Twitter. However, they only experiment with CNNs and some feature embeddings, such as one-hot encoded character n-gram vectors and word embeddings. According to their results, they outperform the baseline in terms of precision and F1-score but not on recall. Similarly,~\cite{park2017one} also use CNNs with character- and word-level inputs for the same task. However, they investigate two different cases; performing the classification for all three labels (`none,' `sexist,' and `racist') at once, or beginning with the detection of `abusive language' and then further distinguish between `sexist' or `racist.' Their results show that, in general, the two cases can have equally good performance. Their deep learning model, though, does not seem to perform as well as traditional machine learning algorithms when it comes to the two-step approach. All previously mentioned works experiment with the dataset that was published in~\cite{waseem2016hateful} and we also use it to compare the results.

Literature on the topic of hate detection on Twitter using deep learning has been sparse. However, there are some works addressing this problem in different online platforms. For example, the study in~\cite{nobata2016abusive} deals with Yahoo news comments which have been annotated as abusive or not, by trained Yahoo employees. More specifically, they employ several datasets from comments found on Yahoo! Finance and News. Most of them were annotated by employees of the company, one was crowdsourced using Amazon's Mechanical Turk and one was provided from~\cite{djuric2015hate}. In the work of~\cite{djuric2015hate}, they also classify hate speech on Yahoo comments, using the continuous BOW neural language model to train word and comment representations into `paragraph embeddings' (named $paragraph2vec$). They classify between hate and clean comments using logistic regression. 

Nobata et al.~\cite{nobata2016abusive} compare directly their research with~\cite{djuric2015hate}, as their works have many similarities. Except from working on the same task and domain, they also employ similar features, i.e., comment embeddings (named $comment2vec$). However, they treat these embeddings differently, without using deep learning based models. Except from the embeddings, they also construct a number of other features, all derived from the text of the comment. For the classification task they use the Vowpal Wabbit's regression model,\footnote{https://github.com/JohnLangford/vowpal\_wabbit} which generally works well with NLP features. According to their results, they outperform~\cite{djuric2015hate} by $0.10$ in AUC and achieve $82.6\%$ F-score.

\smallskip
\textbf{Contributions.}
To the best of our knowledge, the present work is the first to study in depth the application of deep learning on the detection of abuse in all its forms with a unified architecture.
Departing from the previous works that use mostly textual  or custom, task-specific, features,  we  design a neural network architecture that is able to digest all available input (both text and numerical metadata).
Furthermore, training the proposed multi-input network is not straightforward. We introduce an interleaved approach that  has only been adapted from image recommendation systems~\cite{hidasi2016parallel}.
As far as we know, we are the first to experiment with this architecture on classification of text.

To sum up, our work is the first proposed unified solution for detecting a diverse set of abusive behaviors on platforms like Twitter.
The results of our proposed architecture improve over all the state-of-the-art methods, and this is the case across all types of abusive behavior.

\section{Deep Learning Architectures}\label{sec:architecture}

Our overarching goal is to produce a classifier that can detect nuanced forms of abusive behavior. There is a host of literature that tackles this problem and uses a variety of approaches to do so. A key takeaway from the majority of this work is that building a model based on text is outperformed by those that additionally take domain specific features into account. Unfortunately, this is a cumbersome process, with slightly different problems and data sources requiring specially constructed models using different architectures. Ideally, we would prefer to have a \emph{single} model/architecture that incorporates \emph{domain specific} metadata, as well as text content and is performant on a large number of abusive content detection tasks.
To that end, we present a unified classification model for abusive behavior. Our approach is treating i) raw text, and ii) domain specific metadata, separately at first, and later combining them into a single model. 

In the remainder of this section, we provide details on how our final network is built from its component parts, paying particular attention to the specifics of how to train such a multi-input model. Firstly, we present the two individual classifiers that we later fuse: the text and the metadata classifier. Later, we discuss how we can combine them, as well as the different ways of training such a multi-path network.

\begin{figure}[t]
	\centering
	\includegraphics[width=0.9\columnwidth]{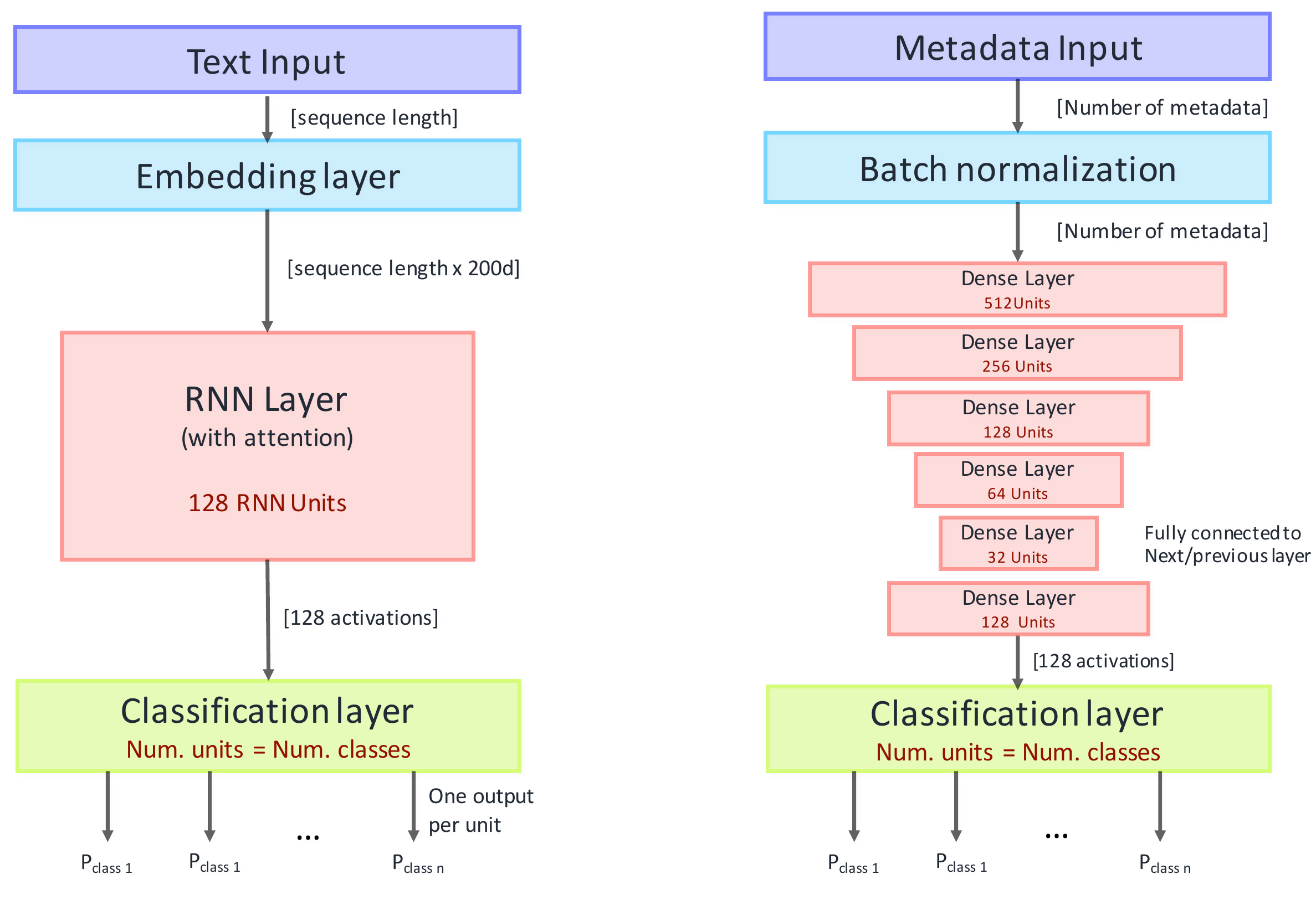}  
	\caption{The individual classifiers that are the basis of the combined model. Left: the text-only classifier, right is the metadata-only classifier.}
	\label{res:single_models}
	\vspace{-0.5cm}
\end{figure}

\subsection{Text Classification Network} 
\label{classifier:text}
This classifier only considers the raw text as input. There are several choices for the class of neural network to base our classifier off. We use Recurrent Neural Networks (RNN) since they have proven successful at understanding sequences of words and interpreting their meaning. We experimented with both character- and word-level RNNs, with the latter to be the most performant across all our datasets.

\smallskip
\noindent\textbf{Text preprocessing:}
Before feeding any text to the network, we need to transform each sample to a sequence of words. As neural networks are trained in mini-batches, every sample in a batch must have the same sequence length (number of words). Tweets containing more words than the sequence length are trimmed, whereas tweets with less words are left-padded with zeros (the model learns they carry no information). Ideally, we want to setup a sequence length that is large enough to contain most text from the samples, but avoids outliers as they waste resources (feeding zeros in the network). Therefore, we take the 95th percentile of length of tweets (with respect to the number of words) in the input corpus as the optimal sequence length. For tweets, this results in sequences of 30 words (i.e., 5\% of tweets that contain more than 30 words are truncated). Sequences with less than this length are padded with zeros. We additionally remove any words that appear only once in the corpus, as they are most likely typos and can result in over-fitting. Once preprocessed, the input text is fed to the network for learning.

\smallskip
\noindent\textbf{Word embedding layer:}
The first layer of the network performs a word embedding, which maps each word to a high-dimensional (typically $25-300$ dimensions) vector. Word embedding has proved to be a highly effective technique for text classification tasks, and additionally reduces the number of training samples required to reach good performance. We settled on using pre-trained word embeddings from GloVe \cite{pennington2014glove}, which is constructed on more than 2 billion tweets. We choose the highest dimension embeddings ($200$) available, as these produce the best results across all abusive behaviors investigated. If a word is not found in the GloVe dataset, we initialize a vector of random weights, which the word embedding layer eventually learns from the input data.

\smallskip
\noindent\textbf{Recurrent layer:} 
The next layer is an RNN with $128$ units (neurons). As mentioned previously, RNNs learn sequences of words by updating an internal state. After experimenting with several choices for the RNN architecture (Gated Recurrent Unit or GRUs, Long Short-Term Memory or LSTMs, and Bidirectional RNNs), we find that due to the rather small sequences of length in social media (typically less than $100$ words per post, just $30$ for Twitter), simple GRUs are performing as well as more complex units. To avoid over-fitting we use a recurrent dropout with $p=0.5$ (i.e., individual neurons were available for activation with probability $0.5$), as it empirically provided the best results across all studied behaviors. Finally, an attention layer~\cite{bahdanau2014neural} can be added as it provides a mechanism for the RNN to ``focus'' on individual parts of the text that contain information related to the task. Attention is particularly useful to tackle texts that contain longer sequences of words (e.g., forum posts). Empirically, we find this only helps for texts that exceed $100$ words and, thus, disable it for any classification task that involves tweets. 

\smallskip
\noindent\textbf{Classification layer:} 
Finally, we use a fully connected output layer (a.k.a. Dense layer) with one neuron per class we want to predict, and a softmax activation function to normalize output values between 0 and 1. The output of each neuron at this stage represents the probability of the sample belonging to each respective class.  Note that this is the layer that is sliced off when we fuse the text and metadata models into the final combined classifier.

\subsection{Metadata Network}
\label{classifier:metadata}
The metadata network considers non-sequential data. For example, on Twitter, it might evaluate the number of followers, the location, account age, total number of (posted/favorited/liked) tweets, etc., of a user.

\smallskip
\noindent\textbf{Metadata preprocessing:}  Before feeding the data into the neural network, we need to transform any categorical data into numerical, either via enumeration or one-hot encoding, depending on the particulars of the input. Then, each sample is thus represented as a vector of numerical features. 

\smallskip
\noindent\textbf{Batch normalization layer:} 
Neural network layers work best when the input data have zero mean and unit variance, as it enables faster learning and higher overall accuracy. Thus, we pass the data through a Batch Normalization layer that takes care of this transformation at each batch.

\smallskip
\noindent\textbf{Dense layers:} 
We use a simple network of several fully connected (dense) layers to learn the metadata. We design our network so that a bottleneck is formed. Such a bottleneck has been shown to result in automatic construction of high-level features~\cite{he2016deep, tishby2015deep}. In our implementation, we experimented with multiple architectures and we ended up using 5 layers of size 512, 245, 128, 64, 32, which provide good results across all studied behaviors. On top of this layer, we add an additional (6th) layer which ensures that this network has the same dimensionality as the text-only network; this ends up enhancing performance when we fuse the two networks. Finally, we use $tanh$ as activation function, since it works better with standardized numerical data.

\smallskip
\noindent\textbf{Classification layer:} 
As with the test only network, we use one neuron per class with softmax activation.

\subsection{Combining the Two Classification Paths}
\begin{figure}[t]
	\centering
	\includegraphics[width=0.9\columnwidth]{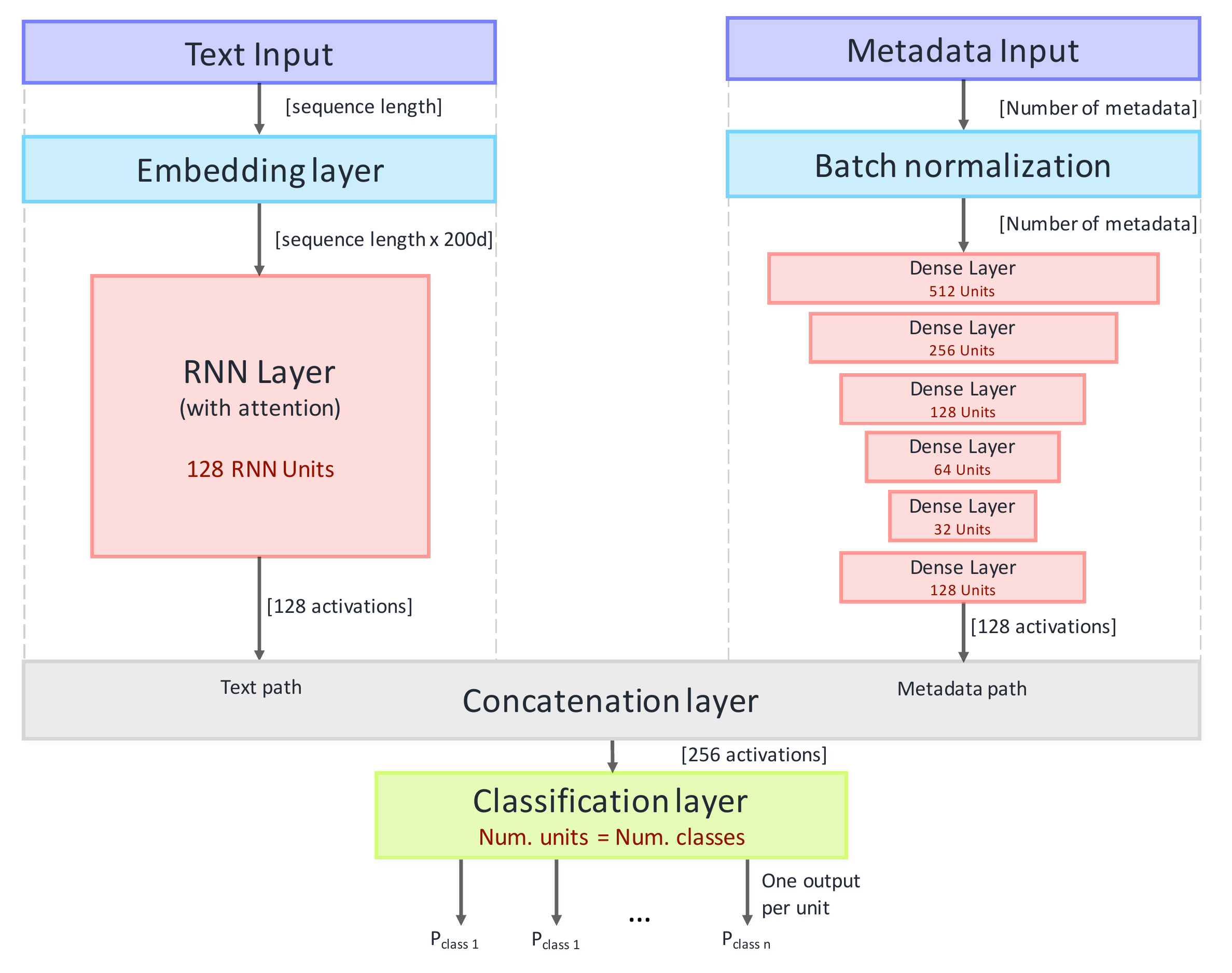}  
	\caption{The combined classifier. The output of the two individual paths are concatenated and a classification layer is added over the merged data.}
	\label{res:combined_model}
	\vspace{-1em}
\end{figure}

The two classifiers presented above can handle individually either the raw text or the metadata. To build a multi-input classifier we need to combine these two paths. There are two possible ways to perform such a task: i) just use the output of the two classifiers (probabilities of belonging to a given class) as input of a new classifier, or ii) combine the two classifiers on the previous layer that represents the automatically constructed features (as shown in Figure~\ref{res:combined_model}). 

Instead of combining an ensemble of pre-trained classifiers, neural networks allow us to create arbitrary combinations of layers and construct complex architectures that resemble graphs. Therefore, instead of training and then combining two separate classifiers, we can design from the beginning a single architecture that combines both paths before their inputs are squashed into  classification probabilities (Figure~\ref{res:combined_model}). Therefore, we concatenate the text and metadata networks at their penultimate layer: i)~the text path where sequences of raw text are input and 128 activations are produced (one for each RNN unit) and ii)~the metadata path where each input  produces 128 activations. We can think of this architecture as merging together 128 automatically constructed features from each input and then attempting the final classification task based on this vector. 

Contrary to traditional machine learning, this architecture allows us to mix a diverse set of data (sequences of text and discrete metadata) without having to explicitly construct the text features (e.g., TF-IDF vectors). Furthermore, we utilize the power of word embedding that have been pre-trained on much larger datasets. 

\subsection{Training the Combined Network}
\label{training}
While the combined architecture is straightforward, just training the whole network at once is not the most optimal way. In fact, there are several ways to train the combined network: below, we list some of the possible ways and on the Evaluation Section we compare their performance. 

\smallskip
\noindent\textbf{Training the entire network at once (Naive Training):}
The simplest approach is to train the entire network at once; i.e., to treat it as a single classification network with two inputs. However, the performance we achieve from this training technique is suboptimal: the two paths have different convergence rates (i.e., one of the paths might converge faster, and thus dominate any subsequent training epochs). Furthermore, standard backpropagation across the whole network can also induce unpredictable interactions as we allow the weights to be modified at both paths simultaneously.

\smallskip
\noindent\textbf{Transfer learning:}
We can avoid this problem by \emph{pre-training} the two paths separately and only afterwards join them together to construct the architecture of Figure~\ref{res:combined_model}.
This involves a number of steps:
\begin{enumerate}
	\item Pre-train separately the text and the metadata classifiers.
	\item We \emph{remove} the classification layer of each classifier, effectively exposing the activations of their penultimate layer. We treat these as the features that the two pre-trained networks have constructed based on their training. 
	\item We freeze the weights of both networks, so no further re-training of their weights is possible.
	\item We add a concatenation layer and a classification layer, effectively transforming the separate models of Figure~\ref{res:single_models} into the architecture of the combined model (Figure~\ref{res:combined_model}). 
	\item We train again the combined model. Only the final layer's weights are trainable. 
\end{enumerate}

\noindent Note: this model resembles an ensemble but with a key difference: the input is not the final class probabilities ($num_{class}$+$num_{class}$ features) but features learned by the previous layer of the pre-trained models (128+128 features).

\smallskip
\noindent\textbf{Transfer learning with fine tuning (FT):}
This approach is the same as above, except we do not freeze the weights on the original networks. The practical result of this is that our pre-training serves only to initialize the weights, which the fused network can later adapt when we merge the two paths.

\smallskip
\noindent\textbf{Combined learning with interleaving (Interleaved):}
As discussed, standard training of the whole network at once may lead to poor performance due to the interaction of updating both data paths at once. The  training approaches presented in the previous sections try to mitigate this problem by training the two paths separately and then concatenating the two pre-trained models together. Instead, here we introduce a way that allows us to \emph{train the full network simultaneously while mitigating the aforementioned drawbacks}. In fact, we demonstrate later that this approach achieves the best results in all four datasets.

To do this, we can design our training in a way that, at each mini-batch, data flow through the whole network, but only one of the paths is updated. To do so, we train the two paths in an alternating fashion. For example, on even-numbered mini-batches the gradient descent only updates the text path whereas on odd-numbered  batches the metadata ones. Finally, between epochs we also alternate the paths so both paths get a chance to observe the whole dataset.

To implement this interleaved approach (e.g., in Keras), we initialize two identical models $A$ and $B$ with the architecture shown in Figure~\ref{res:combined_model}. However, before compiling the models, we introduce a single difference: the text-path of $A$ is defined as non-trainable (`frozen') and, similarly, the metadata-path on $B$ is also `frozen.' During training, at each mini-batch we alternate these models:
\begin{itemize}
	\item If $(batch\_number + epoch\_number)$ is even, then use model A (else, use B). Notice, the input will pass through both paths, however, the gradient will only update the weights of a single path.
	\item Copy the newly updated weights to the unused model. Now both models have equal weights.
	\item Repeat for next mini-batch.
\end{itemize}
At the end of this process, we have two identical models, each trained one-path-at-a-time. Our empirical results shown that mini-batches of 64 to 512 samples perform similar on all datasets and we chose 512 as it speeds up training.

This results in a more optimal, balanced network as the gradient is only able to change one path at a time, thus avoiding unwanted interactions. At the same time, the loss function is calculated over the whole, combined, network (notice that the input does pass through the whole network).

The interleaving architecture, originally introduced in \cite{hidasi2016parallel}, used parallel training on two recurrent neural networks for multimedia and textual features, and applied for video and product recommendations. However, our work is the first one to introduce it on text classification.

\section{Dataset} \label{sec:data}

In this section, we first describe the features we extract. Then we analyze the datasets used in the experiments and how they fit the scope of our analysis. Table~\ref{tab:datasets} summarizes the basic properties (e.g., number of tweets, involved users) and the available metadata per dataset.

\setlength\tabcolsep{1.5pt}
\begin{table}
	\centering
	\small
	\begin{tabular}{|l|c|c|c|c|c|c|c|}
		\hline
		Dataset & Tweets & Classes & Users & WV & TF & UF & NF \\
		\hline\hline
		\multirow{3}{*}{\dswebsci}
		& \multirow{3}{*}{6091} & 8.5\% Bully & \multirow{3}{*}{891} & \multirow{3}{*}{\checkmark} 
		& \multirow{3}{*}{\checkmark} & \multirow{3}{*}{\checkmark} & \multirow{3}{*}{\checkmark}\\	
		& & 5.5\% Aggressive & & & & & \\ & & 86\% Normal & & & & & \\ \hline
		\multirow{3}{*}{\dsnaacl}
		& \multirow{3}{*}{16059} & 12\% Racism & \multirow{3}{*}{1236} & \multirow{3}{*}{\checkmark}
		& \multirow{3}{*}{\checkmark} & \multirow{3}{*}{\checkmark} & \multirow{3}{*}{}\\
		& & 20\% Sexism & & & & & \\ & & 68\% None & & & & & \\ \hline
		\multirow{3}{*}{\dsicwsm}
		& \multirow{3}{*}{24783} & 6\% Hate & \multirow{3}{*}{} & \multirow{3}{*}{\checkmark} 
		& \multirow{3}{*}{\checkmark} & \multirow{3}{*}{} & \multirow{3}{*}{}\\
		& & 77\% Offensive & & & & & \\ & & 17\% Neither & & & & & \\ \hline
		\multirow{2}{*}{\dswsdm}
		& \multirow{2}{*}{61075} & 10.5\% Sarcastic & \multirow{2}{*}{60255} & \multirow{3}{*}{\checkmark} 
		& \multirow{3}{*}{\checkmark} & \multirow{3}{*}{\checkmark} & \multirow{3}{*}{}\\
		& & 89.5\% Normal & & & & & \\ \hline 	
	\end{tabular}
	\caption{A descriptive analysis of the datasets with information about the number of tweets and users, the distribution of the classes and whether or not we have the correspondent word vector (WV), tweet- (TF), user- (UF) and network-based (NF) metadata.}
	\label{tab:datasets}
	\vspace{-0.5cm}
\end{table}

\subsection{Feature Extraction} \label{sec:featuredescription}

\smallskip
\noindent\textbf{Word Vectors} (\textbf{WV}): are representations of words into a vector space (word2vec). As explained earlier, and using the GloVe method \cite{pennington2014glove}, the words from tweets are mapped, or embedded, into a high-dimensional vector of 200 dimensions.

\smallskip
\noindent\textbf{Metadata Features:}
Similar to the work presented in~\cite{chatzakou2017mean} a set of metadata is considered, either tweet-, user-, or network-based, since they have been proven effective for a similar task. More specifically, these metadata are of three general categories:

\smallskip
\noindent\textbf{Tweet-based (\textbf{TF})}: some common and basic textual data are considered, frequently used on Twitter; namely the amount of hashtags and mentions of other users; how many emoticons exist in the tweet; how many words there are with uppercase letters only; the amount of URLs included. Moreover, tweets' sentiment (i.e., positive/negative score), specific emotions (i.e., anger, disgust, fear, joy, sadness, and surprise), and offensiveness scores are considered.\\
\smallskip
\noindent\textbf{User-based (\textbf{UF})}: for the author level, we extract a few basic metadata regarding his popularity (i.e., number of followers/friends). Also, we consider his activity on Twitter based on the number of posted and favorited tweets, the subscribed lists, and the age of his account.\\
\smallskip
\noindent\textbf{Network-based (\textbf{NF})}: we analyze a user's network by considering his followers (i.e., someone who follows a user) and friends (i.e., someone who is followed by a user). Based on \cite{chatzakou2017mean}, the considered metadata indicate a user' popularity (i.e., the number of followers and friends, and the ratio of such measures), the extent to which a user tends to reciprocate the follower connections he receives, the power difference between a user and his mentions, the user's position in his network (i.e., hub, authority, eigenvector and closeness centrality), as well as a user's tendency to cluster with others (i.e., clustering coefficient).

\subsection{\dswebsci Dataset}

The first dataset is provided by~\cite{chatzakou2017mean} and was collected for the purpose of detecting two instances of abusive behavior on Twitter: cyberbullying and cyber-aggression. In addition to a baseline, the authors collected a set of tweets between June and August 2016, using snowball sampling around the GamerGate controversy, which is known to have produced many instances of cyberbullying and cyber-aggression. The $9,484$ tweets were grouped into $1,303$ per user ``batches'' and labeled via crowdsourced workers into one of four categories: 1)~bullying, 2)~aggressive, 3)~spam, or 4)~normal. The authors are careful to differentiate between aggressive and bullying behavior. An aggressor was defined as ``someone who posts at least one tweet or retweet with negative meaning, with the intent to harm or insult other users'' and a bully was defined as ``someone who posts multiple tweets or retweets with negative meaning for the same topic and in a repeated fashion, with the intent to harm or insult other users.'' The aggressive and bullying labels make up about 8\% of the dataset, spam makes up about 1/3, with the remainder normal. For our purposes, we remove the batches labeled as spam, as they can be handled with more specialized techniques~\cite{chatzakou2017mean}. We note that the authors were focused on identifying bullying and aggressive \emph{users}, but we are interested in classifying individual tweets and thus, we break up each batch into individual tweets, each labeled with whatever label their batch was given.
In addition to the vector representations (WV), this dataset includes all types of metadata that we use in the metadata classifier, as previously described (i.e., TF, UF, and NF).

\subsection{\dsnaacl Dataset}

The second dataset, provided by \cite{waseem2016hateful}, is focused on racism and sexism. Collected over a two-month period, the authors manually searched for common hateful terms targeting groups, e.g., ethnicity, sexual orientation, gender, religion, etc. The search results were narrowed down to a set of users that seemed to espouse a lot of racist and sexist views. After collection, the data were preprocessed to remove Twitter specific content (e.g., retweets and mentions), punctuation, and all stop words \emph{except} ``not.''

The tweets were labeled as racist or sexist according to a set of criteria: 1)~if the tweet attacks, criticizes, or seeks to silence a minority, 2)~if it promotes hate speech or violence, or 3)~if there is use of sexist or racial slurs. Data were manually annotated (not via crowdsourcing) and resulted in $2k$ racist and $3k$ sexist tweets, out of $16k$ total. This dataset is a good benchmark for the present work as it has been used by several similar studies, e.g.,~\cite{badjatiya2017deep, gamback2017using, park2017one}.
When working with this dataset, except from the word vectors (WV), we also employ both TF and UF metadata. However, we do not use network-related metadata (NF), due to time limitations (it takes a significant amount of computation and network effort to crawl Twitter users' profiles with nowadays' Twitter API rate limits).

\subsection{\dsicwsm Dataset}

Tweets characterized as hateful, offensive, or neither are provided by \cite{davidson2017automated}. Here, hate speech is defined as language that is used to express hatred, insult, or to humiliate a targeted group or its members. Offensive language is less clearly defined as speech that uses offensive words, but does not necessarily have offensive meaning. Thus, this dataset makes the distinction that offensive language can be used in context that is not necessarily hateful.
Of $80$ million tweets they collected, a $25k$ were labeled by crowdsourced workers, with a resulting intercoder-agreement score of $92\%$. 77\% of the tweets were labeled as offensive, with only $6\%$ labeled as hateful. The authors only made the text of the tweets available, and so we have no metadata to use in our evaluation, other than WV.

\subsection{\dswsdm Dataset}

In the dataset by \cite{rajadesingan2015sarcasm}, tweets are characterized as sarcastic or non-sarcastic. Sarcasm, in this work, is defined as `a way of using words that are the opposite of what you mean in order to be unpleasant to somebody or to make fun of them.' In some online settings such as Twitter or Facebook, with not enough context on the topic of discussion or interest to be civil, sarcasm can be considered impolite and even aggressive behavior. Though this dataset is slightly different from the rest, considering the task at hand, we believe it can bring a useful dimension to the plurality and complexity of abusive behavior, and can inform our methodology on detecting such language.

The data collection was conducted based on self-described users' annotations. Specifically, the authors collected only tweets that contained the hashtags \#sarcasm and \#not. Then, they filtered out tweets that did not contain the aforementioned hashtags at the end of them, in order to eliminate tweets that referred to sarcasm but they were not sarcastic. Moreover, they removed the non-english tweets, retweets, tweets with less than three words, as well as tweets that contained mentions or URLs (due to computational complexity). The final dataset consists of almost $91k$ tweets, where 10\% are sarcastic. Since not all data were still publicly available through Twitter API, we ended up with $60k$ - preserving the portion of sarcastic tweets. Finally, during the classification, we removed the \#sarcasm and \#not hashtags.

Similarly with the \dsnaacl dataset, here also we employ WV, TF and UF attributes, but no NF. For our experiments, we use the original highly imbalanced dataset (even though the authors report attempts of data balancing in order to improve the classification performance) since it adapts better to real cases. Hence, we compare our classification performance with the imbalanced results of the baseline.

\section{Evaluation}\label{sec:evaluation}

In this section, we describe in detail our experimental setup and results while testing the performance of our method on the different datasets used. All results shown here are based on 10-fold cross validation.  

\subsection{Experimental Setup}

For our implementation we use Keras\footnote{https://keras.io/} with Theano\footnote{http://deeplearning.net/software/theano/} as back-end for the deep learning models implementation. We use the functional API to implement our multi-input single-output model. Finally, we run the experiments on a server that is equipped with three Tesla K40c GPUs.

In terms of training, we use \emph{categorical cross-entropy} as loss function and \emph{Adam} as the optimization function. A maximum of 100 epochs is allowed, but we also employ a separate validation set to perform \emph{early stopping}: training is interrupted if the validation loss did not drop in 10 consecutive epochs and the weights of the best epoch are restored.

It is important to notice that the same model (with the same architecture, number of layers, number of units and parameters) is used for all datasets, as we want to demonstrate the performance of this architecture across different tasks. The performance of the algorithm might increase even further if the parameters are tuned specifically for each task (e.g., using a larger network when there are more training samples). Overall, the model, excluding the pre-trained word embeddings, contains approximately 250,000 trainable parameters (i.e., weights).

Finally, except from each dataset's state-of-the-art, we also compare our results with a basic Naive Bayes model, using the TF-IDF weights for each tweet. For this baseline, we only use the raw text. First, we perform some basic preprocessing of the data; we convert all characters to lowercase and remove all stop words for 14 frequently spoken languages, as well as some twitter-specific stop words. Finally, we tokenize the tweet based on some Twitter-specific markers (hashtags, URLs, and mentions) and punctuation. Afterwards, we experiment with both Porter and Snowball stemmers, lemmatization, keeping the most frequent words and the combinations of all the previously mentioned. We find that the most efficient step is keeping only the most frequent words. We also experiment on the amount of frequent words we need to keep and find that the best results are yielded using the top $10k$ words. 

\vspace{-0.3cm}
\subsection{Experimental Results}

In this section we present the classification performance of the proposed methodology on the four datasets. Next, we  examine which are the inputs that contribute the most. Finally, we  discuss about how the different training strategies affect the classification performance. 

\begin{table}[t]
	\centering
	\small
	\begin{tabular}{lp{0.6cm}p{0.6cm}p{0.6cm}p{0.6cm}p{0.6cm}}
		\hline
		& AUC & Acc. & Prec. & Rec. & F1 \\ \hline\hline
		\multicolumn{6}{|c|}{\dswebsci Dataset (3 classes)} \\ \hline
		DL-Baseline Naive Bayes & 0.73 & 0.88 & 0.88 & 0.88 & 0.88 \\
		Chatzakou et al. 2017		& 0.91 & 0.91 & 0.90 & 0.92 & 0.91 \\
		DL-Metadata only & 0.93 & 0.88 & 0.91 & 0.88 & 0.89 \\
		DL-Text only & 0.92 & 0.89 & 0.91 & 0.89 & 0.89 \\
		DL-Text \& Metadata (Naive Train.) & 0.94 & 0.89 & 0.90 & 0.90 & 0.90 \\
		DL-Text \& Metadata (Tran. Lear.) & 0.95 & 0.90 & 0.92 & 0.90 & 0.90 \\
		DL-Text \& Metadata (Tran. Lear. FT) & 0.95 & 0.90 & 0.91 & 0.90 & 0.91 \\
		DL-Text \& Metadata (Interleaved) & 0.96 & 0.92 & 0.93 & 0.92 & 0.93 \\ \hline
		\multicolumn{6}{|c|}{\dsnaacl Dataset} \\ \hline
		Baseline Naive Bayes & 0.79 & 0.81 & 0.81 & 0.81 & 0.81 \\
		Waseem and Hovy 2016	& - & - & 0.74 & 0.73 & 0.78 \\
		DL-Metadata only & 0.91 & 0.74 & 0.81 & 0.74 & 0.76 \\
		DL-Text only & 0.93 & 0.83 & 0.84 & 0.83 & 0.83 \\
		DL-Text \& Metadata (Naive Train.) & 0.93 & 0.85 & 0.86 & 0.86 & 0.86 \\
		DL-Text \& Metadata (Tran. Lear.) & 0.95 & 0.85 & 0.86 & 0.85 & 0.85 \\
		DL-Text \& Metadata (Tran. Lear. FT) & 0.95 & 0.86 & 0.87 & 0.86 & 0.86 \\
		DL-Text \& Metadata (Interleaved) & 0.96 & 0.87 & 0.88 & 0.87 & 0.87 \\ \hline    
		\multicolumn{6}{|c|}{\dsicwsm Dataset} \\ \hline
		Baseline Naive Bayes & 0.71 & 0.87 & 0.84 & 0.87 & 0.85 \\
		Davidson et al. 2017	& 0.87 & 0.89 & 0.91 & 0.9 & 0.9 \\
		DL-Metadata only & 0.75 & 0.61 & 0.80 & 0.61 & 0.66 \\
		DL-Text only & 0.91 & 0.87 & 0.89 & 0.87 & 0.88 \\
		DL-Text \& Metadata (Naive Train.) & 0.90 & 0.87 & 0.89 & 0.87 & 0.88 \\
		DL-Text \& Metadata (Tran. Lear.) & 0.91 & 0.87 & 0.89 & 0.87 & 0.88 \\
		DL-Text \& Metadata (Tran. Lear. FT) & 0.90 & 0.87 & 0.89 & 0.87 & 0.88 \\
		DL-Text \& Metadata (Interleaved) & 0.92 & 0.90 & 0.89 & 0.89 & 0.89 \\ \hline
		\multicolumn{6}{|c|}{\dswsdm Dataset} \\ \hline
		Baseline Naive Bayes & 0.66 & 0.90 & 0.89 & 0.9 & 0.89 \\
		Rajadesingan, Zafarani, and Liu 2015	& 0.7 & 0.93 & - & - & -\\
		DL-Metadata only & 0.96 & 0.92 & 0.94 & 0.92 & 0.92 \\
		DL-Text only & 0.81 & 0.89 & 0.89 & 0.89 & 0.89 \\
		DL-Text \& Metadata (Naive Train.) & 0.97 & 0.96 & 0.96 & 0.96 & 0.96 \\
		DL-Text \& Metadata (Tran. Lear.) & 0.97 & 0.95 & 0.95 & 0.95 & 0.95 \\
		DL-Text \& Metadata (Tran. Lear. FT) & 0.97 & 0.95 & 0.95 & 0.95 & 0.95 \\
		DL-Text \& Metadata (Interleaved) & 0.98 & 0.97 & 0.96 & 0.97 & 0.97 \\ \hline
	\end{tabular}
	\caption{Final results of the baselines and our experiments, for each one of the datasets.}
	\label{tab:results}
	\vspace{-0.5cm}
\end{table}

\smallskip
\noindent\textbf{Training methodology:}
We apply the same model over all 4 datasets and the results of AUC, Accuracy, Precision, Recall, and F1-score are summarized in Table~\ref{tab:results}. Here, we test the training methods discussed earlier to choose the best method to compare with the state-of-art. Firstly, we observe that training with the whole network at once (naive training) results to suboptimal performance (e.g., AUC of 0.94 in the \websci dataset). The reason is that allowing the gradient descent to update both paths simultaneously might result in unwanted interactions between the two. For example, one path might converge faster than the other, dominating in the decisions. In fact, when we examine the standalone classifiers, we observe that the text classifier requires 25-40 epochs to converge whereas the metadata classifier only requires 7-12. By training the whole network together, the metadata side can start overfitting.

A step towards the right direction is to train each path separately, as individual classifiers, and then transfer the constructed features to a new classifier (transfer learning). This methodology slightly improves the results as it reduces the interactions between the two paths. Notice that, due to the fact that most of the network has been already trained, the additional layer converges after just 3-5 epochs. 

Finally, by employing alternate training (interleaved), we further improve the predictive power of the resulting model reaching 0.96 AUC in the \websci. This shows that multi-input models such as this can benefit from alternate training. The reason is that this methodology avoids any interactions that might result when weights are updated simultaneously in both paths. 

\smallskip
\noindent\textbf{Classification performance:}
Across all datasets and abusive behaviors, the proposed classifier (Interleaved) outperforms both the baseline and the state-of-art, as reported in recent publications. This happens for two reasons. First, the proposed approach that combines the raw text and the metadata achieves notably higher performance when compared to the ones using a single set of attributes, as it takes advantage of the additional information from the users' profile and network. This is consistent across all four datasets. Second, the word embeddings allow us to transfer features that were constructed over a billion of tweets, and it enables us to model complex tasks with fewer samples (such as this one).

Looking at the four datasets, on the \websci dataset we observe that using a single set of attributes (text or metadata) we achieve better AUC (0.92 or 0.93, respectively) but worse accuracy (0.89 or 0.88, respectively) than the method proposed in the state-of-the-art (AUC 0.91, accuracy 0.91)~\cite{chatzakou2017mean}. However, the interleaved training of our model substantially outperforms the baseline (AUC 0.96, accuracy 0.92). 
Having said that, we need to mention here that the comparison with this dataset is not direct. The results reported on~\cite{chatzakou2017mean} are on user-level, while ours are on tweet-level. Therefore, while we can get an understanding on how well their data can be classified with our algorithm, we cannot parallelize the two cases. Nevertheless, the results we achieve on tweet-level are very high, which shows that our model distinguishes very well between the classes, regardless of the comparison.

On the \naacl and \wsdm datasets, we largely outperform the previous results, as these works did not consider metadata. For example, in \wsdm we reach an AUC of 0.98 compared to 0.7 of the existing methodology, as in this case the text is not carrying significant information to detect if a tweet is sarcastic. However, the remaining metadata (user, network, sentiment, tweet-level metadata) do reveal such information. In the case of \naacl dataset, there was no AUC reported in the original publication, but the combination of text and metadata reached a precision and recall of 0.87-0.88, when compared to 0.73-0.74 of the baseline.

Finally, in the \icwsm dataset the interleaved model is able to reach an AUC of 0.92. It is the only case where the precision and recall is similar to baseline, the one presented in ~\cite{davidson2017automated}, but our AUC and accuracy scores still outperform this baseline. This is due to the fact that we could not find any user or network metadata related to this dataset and, therefore, our classifier is only using the raw text and the tweet-based metadata.

\smallskip
\noindent\textbf{Metadata importance:}
As described earlier, we use a number of metadata features extracted from the tweets and their authors, namely tweet-based, user-based, and network-based. These metadata play an important role on the improvement of the performance. When combined with the raw text, they substantially increase all the metrics. However, not all of them have the same impact on the performance (some are more essential than others). In order to determine how each one of these metadata affects our model, we experiment with the \websci dataset and calculate their importance. The results on the AUC are presented in Table~\ref{tab:importance}. We chose this dataset as we have all groups of metadata available (user, tweet, network, text) and, therefore, it is possible to examine how each of them contributes to the model. The results for other datasets are following similar trends and are omitted due to space limitations.

\begin{table}
	\centering
	\small
	\begin{tabular}{llc}
		\hline
		Metadata Features	&	Acronym	& AUC\\ \hline  
		Network Only		&		NF	& 0.641\\
		Tweet Only		&		TF	& 0.799\\
		User Only			&		UF	& 0.806\\
		User \& Tweet		&	UF+TF	& 0.887\\
		Network \& Tweet	&	NF+TF	& 0.908\\
		Text Only			&	WV		& 0.915\\
		User \& Network	&	UF+NF	& 0.915\\
		All-metadata Only	&	TF+UF+NF& 0.923\\
		Text \& Tweet		&	WV+TF	& 0.930\\
		Text \& Network		&	WV+NF	& 0.931\\
		Text \& User \& Tweet	&	WV+UF+TF	&	0.933\\
		Text \& Network \& Tweet	&	WV+NF+TF	&	0.936\\
		Text \& User			&	WV+UF		&	0.938\\
		Text \& User \& Network	&	WV+UF+NF	&	0.955\\
		All 					&	WV+TF+UF+NF& 0.961\\ \hline
	\end{tabular}
	\caption{Metadata Importance. The values are obtained using the \dswebsci dataset.}
	\label{tab:importance}
	\vspace{-0.5cm}
\end{table}

Firstly, by examining individual metadata, we observe that models which are built with individual metadata result in the poorest performance. For example, network-level metadata are the least descriptive (AUC of just 0.64) whereas tweet- and user-based are slightly better (AUC ~0.8). By far, raw text is the best feature for this task, as it can lead to a model with particularly higher AUC of 0.91.

Furthermore, combining two or more metadata classes together (user, network, tweet) does increase performance. This indicates that the information provided is not overlapping and it all adds to better performance. When all the metadata are combined, we reach an AUC of 0.923 which is higher than any other metadata combination. Moreover, using all metadata does result in better classification when compared to just using the text, showing that these metadata carry at least as much predictive power as text does.

Nevertheless, the strongest models were built when text is combined with metadata showing how much the raw text contributes in this classification task. For example, just combining text with user-level metadata is enough to reach an impressive performance (0.94 AUC). Adding network data bumps the performance to 0.955. 

Finally, the best performance is reached when all attributes are used. This also demonstrates the fact that the metadata information does not overlap the information that can be extracted from raw text, and this is why the proposed model can be quite powerful and outperforms the state-of-art models for these tasks.

\section{Generalizing to Other Platforms: Toxic Behavior in Online Gaming}

Though in this work we primarily focus on Twitter to demonstrate how our unified approach works with the same set of features, the same methodology can be applied to other domains \emph{with no modifications}. To demonstrate this, we run the same architecture over a dataset from a completely different domain. We acquired the dataset from the authors of~\cite{toxic_gaming} who built a classifier to detect toxic behavior in an online video game. Their dataset is collected from a crowdsourced system that presents reviewers with instances of millions of matches, where toxic behavior is potentially exhibited. The match data include a variety of details such as the full in-game chat logs, players' in-game performance, the most common reason the match was reported for, the outcome of the match, etc. Matches are labeled for either \emph{pardon} or \emph{punish} by a jury of other players who cast votes in either direction.

From the 1 million individual matches provided to us, we extracted a set of features. Like the datasets used earlier in this work, we extract the chat logs of each potentially offending player. We also extract a set of domain specific metadata features (e.g., features that describe the offenders' performance, as well as the performance of other players, the outcome of the game, how many reports the match received, the most common report type, etc.). Each match is also labeled with the final decision of the crowdsourced worker (either, pardon or punish).

Even though, at a high level, this dataset is structured similarly to the 4 datasets presented earlier (i.e., it is divided into text based and domain specific categories), there are important differences. First, the text is \emph{much} larger (on average offenders use 2,500 words per match compared to just 30 words per tweet). Next, the domain specific nature of the metadata does not really have an analogue in the Twitter datasets we used. Finally, the language used in the chat logs themselves, while English, is littered with domain specific jargon. Thus, applying our architecture to this dataset makes a strong case for its portability to different domains.

In~\cite{toxic_gaming}, the authors evaluated several sub-tasks, with varying degrees of difficulty. The first was the general problem of predicting whether a player will be pardoned or punished, where their best model had an AUC of 0.80. They also experimented with trying to predict only overwhelming decisions, an easier problem, and achieve AUCs of 0.88 and 0.75 for overwhelming pardon and punish decisions, respectively.

We tackled the more general problem by running the dataset through the model presented in our Architecture Section. We enabled an attention layer to deal with the length of the text, however, no other changes were made to the architecture. While we expected \emph{reasonable} performance, we achieved an accuracy of 0.93 and an AUC of 0.89, beating the performance of even the easiest task presented in~\cite{toxic_gaming}. These results provide a strong indication that our architecture is suitable for finding abusive behavior in a wide variety of domains.

\section{Summary}\label{sec:discussion}

\smallskip
\noindent\textbf{Unified deep learning classifier is possible:} In this work, we built and applied the exact same deep learning model architecture in all four datasets and demonstrated that it can efficiently handle each type of abusive behavior. While fine-tuning the classifier parameters for each dataset can squeeze some more performance, the proposed methodology does beat the current state of art in each behavior detection.

\smallskip
\noindent\textbf{All inputs help:} We demonstrated how each of the attributes (text, user, network, tweet) contribute in each task, i.e., identification of specific type of abusive behavior. Our proposed architecture can seamlessly combine this input into a single classification model, without particular tuning.

\smallskip
\noindent\textbf{Training methodology:} Training a multi-input network is not straightforward. We introduced a methodology that alternates training between the two input paths to further increase performance in all datasets tested. We compared the proposed training paradigm with various other possible training methodologies (ensemble, feature transfer, concurrent training) and show that it can substantially outperform them.

\smallskip
\noindent\textbf{Flexible to other data:} In this paper, we show the ability of our approach to combine two different paths: text and metadata. However, one can simply concatenate more input paths to the architecture. For example, in an image classification problem, a CNN-based network can be used to extract image features and it could be joined with text information (tags and user comments) and image metadata (time and location taken, how many pictures the user has taken, the uploader's social network, etc.). Similarly, in an audio classification task, an audio path can be merged with text and metadata. We leave this exploration as future work. 

\smallskip
\noindent\textbf{Generalizing to other platforms:} Finally, we showed that our proposed architecture can be easily applied, in a plug-and-play fashion, to detect abusive behavior in other online domains beyond Twitter. As an example, we presented results on detecting toxic behavior in an online gaming network, with superior performance over the state of art. We leave further explorations as future work.

\section{Acknowledgements}\label{sec:acknowledgements}

The authors acknowledge research funding from the European Union’s Horizon 2020 research and innovation programme under the Marie Skłodowska-Curie Grant Agreement No 691025.

\small
\bibliography{bibfile}
\bibliographystyle{aaai}
\end{document}